\pdfoutput=1

\documentclass[11pt]{article}

\usepackage[]{acl}

\usepackage{times}
\usepackage{latexsym}
\usepackage{amsmath}

\usepackage[T1]{fontenc}

\usepackage[utf8]{inputenc}

\usepackage{microtype}
\usepackage{graphicx}
\usepackage{hyperref}
\usepackage{inconsolata}

\title{Clustering Document Parts: \\ Detecting and Characterizing Influence Campaigns from Documents}

\author{Zhengxiang Wang \\
  Department of Linguistics and IACS \\  
  Stony Brook University \\
  \texttt{zhengxiang.wang@stonybrook.edu} \\ \And
  Owen Rambow \\
  Department of Linguistics and IACS\\  
  Stony Brook University \\
  \texttt{owen.rambow@stonybrook.edu} \\}

\begin{document}
\maketitle
\thispagestyle{plain}
\pagestyle{plain}
\begin{abstract}

We propose a novel clustering pipeline to detect and characterize influence campaigns from documents. This approach clusters parts of document, detects clusters that likely reflect an influence campaign, and then identifies documents linked to an influence campaign via their association with the high-influence clusters. Our approach outperforms both the direct document-level classification and the direct document-level clustering approach in predicting if a document is part of an influence campaign. We propose various novel techniques to enhance our pipeline, including using an existing event factuality prediction system to obtain document parts, and aggregating multiple clustering experiments to improve the performance of both cluster and document classification. Classifying documents after clustering not only accurately extracts the parts of the documents that are relevant to influence campaigns, but also captures influence campaigns as a coordinated and holistic phenomenon. Our approach makes possible more fine-grained and interpretable characterizations of influence campaigns from documents. 

\end{abstract}

\begin{table*}
\centering
\begin{tabular}{p{1.5cm}p{6cm}p{6cm}}
\hline
\textbf{Media} & \textbf{Positive} & \textbf{Negative}\\
\hline
Twitter & ...Putin \textbf{cleans} up the bioweapons labs installed by the deep state... (44 tks) & ...RT @EmmanuelMacron France strongly \textbf{condemns} Russia's decision to wage war on Ukraine... (19 tks) \\ \hline
Forum & ...a secret NATO \textbf{laboratory} for biological weapons...Biological weapons tests were \textbf{carried} out in the laboratories of this facility... (638 tks)  & ...[NATO] has \textbf{blocked} Ukraine's plan to enter...Item 3: Ukraine \textbf{was} a pawn that the Westerners deliberately sacrificed to strengthen NATO... (703 tks) \\ \hline
News & ...a NATO secret biological \textbf{laboratory} with biological weapons...The biological laboratory under the Azovstal plant in Marioupol in the so-called PIT-404 facility was \textbf{built}...In the laboratories of the facility, tests were \textbf{carried} out to create biological weapons... (1497 tks)  & ...Russia's \textbf{demand} for neutrality...But NATO members \textbf{said} that Ukraine's membership was at best a distant option... [The leader of the Ukrainian separatist region of Lugansk said he could hold a referendum on integration into Russia,] a \textbf{decision} immediately criticized by Kiev...(1152 tks)  \\
\hline
\end{tabular}
\caption{\label{tab:runningExamples}
Two clusters of document parts in three media forms that reflect (positive) or do not reflect (negative) an influence campaign related to the Ukraine bioweapons conspiracy theory. The number of tokens in each document is indicated in the parenthesis following each text. We use ``...'' to highlight and separate the parts of document that are inside the two clusters. The document parts inside the positive cluster, called high-influence cluster in Sec~\ref{sec:pipeline}, present evidence of an influence campaign and the related documents are thus likely to be linked to an influence campaign. Here the document parts are beliefs of the author (see Sec~\ref{sec:experiments}) with the target words highlighted in bold.}
\end{table*}

\section{Introduction \label{sec:introduction}}

Inspired by \citet{Diego2023} and \citet{luceri2023leveraging}, we define an influence campaign as a \emph{coordinated and strategic effort} to shape and manipulate the perceptions of a target audience about certain things or issues over a period of time. It can be organized by an individual, organization, or government for various purposes, such as promoting a specific public image, product, policy, or political narrative.  It can be carries out through various channels, including traditional media and online platforms. Consequently, detecting an influence campaign requires \emph{holistic evaluations} and the use of multiple indicators, such as the social network \citep{mesnards2018detecting}, that point to a collective effort with a shared motive that aims to impact public opinions in a certain way. Accurate and reliable detection typically involves extensive manual verification by domain experts, taking into account both textual and non-textual information \citep{Diego2023}.

In the context of NLP, detecting influence campaigns typically means predicting if an input document is part of an influence campaign \citep{luceri2023leveraging}, i.e., \emph{a binary classification task}. However, this is a different task from capturing the phenomenon of influence campaigns, which naturally is \emph{a clustering problem}, i.e., grouping a collection of documents that reflect an influence campaign. 

\textbf{In practice, the classification task is difficult if not doomed, because by definition, an influence campaign cannot possibly be inferred from a single document.} Consider the examples in Table~\ref{tab:runningExamples}, where the texts in the ``Positive'' column reflect an influence campaign linked to the Ukraine bioweapons conspiracy theory\footnote{\url{https://en.wikipedia.org/wiki/Ukraine_bioweapons_conspiracy_theory}.} and the texts in the ``Negative'' column do not. The only thing that connects the positive texts and distinguishes them from the negative texts is the \emph{shared theme/belief}, expressed by \emph{some parts} of each document (short or long), that there exist US biolabs in Ukraine for the purpose of developing bioweapons. Arguably, any text classifier trained on some specific influence campaign datasets will at best be reduced to detecting some key words expressing the themes of the influence campaigns in the training data; such a classifier will have brittle generalization capacity. Moreover, having a binary classification decision about whether a document reflects an influence campaign neither tells us {\em how} the document reflects an influence campaign, nor does it reveal {\em what the influence campaign is about}. In contrast, if we have a cluster of documents relevant for an influence campaign clustered together, such as those in Table~\ref{tab:runningExamples}, it not only makes it possible to characterize the theme of the influence campaign, but it also makes it much more straightforward to understand why each document is part of an influence campaign: it is because the document, along with other documents in the cluster, contains certain document parts that express an orchestrated theme.


In this paper, we propose a novel text-only \emph{clustering-based} pipeline to help detect and characterize influence campaigns from documents. Unlike the typical document-level classification approach discussed above, the pipeline \emph{predicts influence campaigns directly on the cluster level}, i.e., it predicts whether a cluster of document parts present an influence campaign (a high-influence cluster). From there, the pipeline further predicts whether any document associated with a high-influence cluster is part of the influence campaign \emph{via a dynamic projection procedure}. As a result, \textbf{our pipeline is capable of handling the two aspects of the influence campaign detection task: capturing influence campaigns as a holistic phenomenon, and predicting documents that are part of an influence campaign.} Since influence campaigns are captured by clusters of document parts and the documents predicted to engage in an influence campaign are projected from these clusters, our pipeline enables \emph{fine-grained and interpretable characterizations} of influence campaigns from documents. The specific contributions of this paper are as follows.

\begin{itemize}
    \itemsep-0.15em
    \item We introduce a novel clustering pipeline that detects influence campaigns on both cluster and document levels. This approach significantly outperforms the direct document-level classification approach and the document-level clustering approach.  We do not use lexical features in any of our experiments so as not to overfit to the dataset we use.

    \item We propose a new approach to the classification of documents based on clustering parts of the document.  We show that this approach outperforms clustering documents for our task.  This approach makes possible fine-grained and interpretable characterizations of \emph{what parts of a document lead to the classification of the document}.

    \item We present the very first study to use multi-word text spans expressing certain belief of an entity about the factuality of an event in the input text to extract document parts. We show that for the influence campaign detection task, clustering these text spans can improve the detection performance of influence campaigns from documents, compared to simply clustering sentences.
    
    \item We show that instead of optimizing for a clustering algorithm and its parameters, using an aggregation of algorithms and parameters performs better in our classification task, and provides for more stable results.    
    
\end{itemize}

This paper is structured as follows. We review related works in Sec~\ref{sec:review} and motivate and explain the algorithmic idea underlying our novel clustering pipeline in Sec~\ref{sec:pipeline}. We describe the influence campaign detection dataset we test our pipeline on in Sec~\ref{sec:data} and the experiments in Sec~\ref{sec:experiments}. The results are discussed in Sec~\ref{sec:results}. The paper concludes in Sec~\ref{sec:discussion}. We release our code at \url{https://github.com/jaaack-wang/detect-influence-campaigns}.

\section{Related Work \label{sec:review}}

There have been very few studies in the existing literature that approach influence campaigns in the general sense as we define it in Sec~\ref{sec:introduction}. The influence campaigns studied in most previous research \citep{mesnards2018detecting, luceri2023leveraging, Diego2023} are political influence campaigns, or 
some closely related political influence operations that may be an influence campaign, such as the spreading of mis/dis-information \citep{Ferrara_2017, FIGUEIRA2017817, Rubin2017DeceptionDA, Addawood2019LinguisticCT,Proppy2019, Nogara2022, sakketou-etal-2022-factoid, Malik2023}.

The most common detection method relevant for influence campaigns is the detection of bots in social networks \citep{Davis2016, Badawy2018, mesnards2018detecting, Himelein2021, Hajli2022, Rossetti2022BotsDA}. For text-based influence campaign detection, various NLP methods have been explored. For example, a recent study leverages LLMs \citep{luceri2023leveraging} to predict if a tweet is part of an known influence campaign. Other studies relevant for influence campaigns utilize various sources of linguistic features (e.g., lexicon counts, ngrams, word embedding) to train or fine-tune different models (e.g., BERT, graph neural network, decision tree) with a goal to detect propagandistic, deceptive, or misleading information \citep{Addawood2019LinguisticCT, Proppy2019, sakketou-etal-2022-factoid, Malik2023}. To the best of our knowledge, we are not aware of any study that aims to detect influence campaigns on the cluster level.

Several twitter datasets have been used by recent studies on detecting influence campaigns \citep{mesnards2018detecting, luceri2023leveraging}, such as the 2016 US election dataset \citep{PDI7IN2016}, data from Twitter's Information Operations archive\footnote{E.g., \url{https://blog.twitter.com/en_us/topics/company/2020/2020-election-changes}}, and Russian troll accounts for 2016 US election released by the U.S. Congress \citep{Addawood2019LinguisticCT}. There are also other relevant datasets in other media forms, such as FACTOID \citep{sakketou-etal-2022-factoid} collected from Reddit and Proppy \cite{Proppy2019} collected from news articles. We note that for all of these datasets that come with labels, the labels are typically created on the basis of some simplistic association or assumption. For example, tweets are assumed to be linked to an influence campaign if they come from Russian troll accounts \citep{luceri2023leveraging}. To the best of our knowledge, we do not know of any publicly available influence campaign datasets that contain more than one media type.

\section{Pipeline: The Algorithmic Idea \label{sec:pipeline}}

Given the coordinated nature of influence campaigns, an influence campaign can be thought of as a cluster of documents that spread a certain theme aimed to influence the target audience. Our pipeline follows exactly this intuition and transforms the task of influence campaign detection into one that detects clusters that are highly likely to reflect an influence campaign (i.e., high-influence clusters). Then the next step naturally becomes how to accurately select documents (i.e., high-influence documents) associated with the high-influence clusters that reflect an influence campaign, assuming the clusters may contain some noise or false positives. 

More concretely, our pipeline consists of the following four steps.

\paragraph{Determining document parts} In a pre-processing step, we start out by extracting parts from a document.  In this paper, we experiment with three types of document parts: the multi-word text spans that represent what the author expresses certain belief in (see Sec~\ref{sec:ourApproach}); sentences; and the whole document. 

\paragraph{Clustering parts of documents} Given a set of documents, the pipeline clusters the document parts. Clustering parts of documents not only creates a complex connection network among documents via their semantically related parts, but also presents a general and effective workaround for long document information retrieval using unsupervised clustering algorithms \citep{mekontchou2023information}. 

\paragraph{Classifying high-influence clusters} At training time, the pipeline takes as inputs a collection of documents, each of which is annotated with a binary label: the document is or is not part of an influence campaign. The fact that our pipeline requires annotated documents during training highlights that it is a supervised approach. The concept of high-influence clusters is defined by a user-given threshold $\alpha$, denoting the minimum percentage of document parts in a cluster linked to documents from an influence campaign for the cluser to count as a high-influence cluster. 

Intuitively, $\alpha$ should be set far greater than 0.5 to align with the heuristic that a high-influence cluster should be dominated by document parts from documents that are part of an influence campaign. The assumption here is that parts of documents with a link to an influence campaign are \emph{unlikely} to be clustered together unless they are related to some aspect or surface theme of the influence campaign. Since high-influence clusters may be rare or even absent for a given clustering experiment for which the majority of documents are ``innocent'' and do not reflect an influence campaign, we propose doing multiple clustering experiments and aggregating the resulting clusters together as a way to generate more data to train a classifier for high-influence clusters.  This approach is a novel \emph{data augmentation} technique for cluster-level classification. In this paper, we set $\alpha=0.95$. We set $\alpha \neq 1$ as a trade of the precision and recall for discovering high-influence clusters, since allowing a small error term $1-\alpha$ ($\alpha \neq 1$) in the definition of high-influence clusters facilitates discovery of more high-influence clusters, i.e., a great improvement in recall at a small cost of precision, ultimately leading to a better F1. Note that $\alpha$ is only set and used at the training time. 

At inference time, the pipeline deploys the pre-trained classifier to detect high-influence clusters by predicting the likelihood of a cluster being a high-influence cluster.


\paragraph{Classifying high-influence documents} High-influence documents are documents with connections to high-influence clusters, meaning at least some of their parts occur in at least one high-influence clusters. Formally, we set a threshold $\beta$, a ratio for the number of high-influence clusters, which denotes the minimum number of times parts of a document that must occur in any high-influence clusters to qualify the document as a high-influence document. The threshold $\beta$ is used to control the number of false positives (i.e., documents with no link to an influence campaign occurring in high-influence clusters) introduced by the threshold $\alpha$ set in the previous step. We wish to come up with a module in our current system that predicts an optimal threshold $\beta$ in the future. \\

In summary, we have introduced the notion of ``high-influence clusters'' based on parts of documents and the notion of ``high-influence documents'' based on their association with high-influence clusters. We propose two thresholds ($\alpha$ and $\beta$) to regulate the number of false-positive high-influence documents our system may end up selecting from high-influence clusters. The threshold $\alpha$ is used for training a high-influence cluster classifier, whereas the threshold $\beta$ is used at the time of classifying high-influence documents. We also propose the aggregation of clustering experiments, instead of hard fine-tuning for an optimal clustering experiment, to reliably enhance model performance. 

In what follows, we show that our approach can easily and significantly outperform direct document-level classification, in an apples-to-apples comparison, when it comes to detecting influence campaigns from documents.


\section{Data \label{sec:data}}

We use data collected during a large research program, DARPA INCAS project\footnote{\url{https://www.darpa.mil/program/influence-campaign-awareness-and-sensemaking}}. We expect the data to be made public after the end of the research program. We use this dataset as we are not aware of any other datasets that have expert-verified annotations indicating if a collection of documents contain influence campaigns. 

The data contains four piles of online posts published during January 31 to June 30, 2022. Each pile is a collection of documents in six media forms, namely, Twitter, Forum, News, Blog, Reddit, and Other. Two of the four piles contain documents that engage in an influence campaign that spreads disinformation related to Ukraine bioweapons conspiracy theory
(see Table~\ref{tab:runningExamples}), whereas the other two contain Ukraine-related documents with no links to any known influence campaigns. Lexical search was used to facilitate the collection of the data. Over 99\% of the documents that participate in the bioweapons influence campaign use words like ``biolab'' and ``biological weapons'', but slightly less than 3\% of the documents unrelated to the campaign mention these terms. 
That means that any content-based text classifier, whether rule-based or neural, will overfit this dataset by capturing the related keywords. We avoid training such classifiers, as (1) we are more interested in developing a \emph{potentially general} approach that can both detect and characterize influence campaigns from documents; and (2) content-based text classification for the influence campaign detection task arguably cannot be a general approach nor can it make predictions at beyond the document level to capture the phenomenon of influence campaign.

The majority of these documents are written in French, typically accompanied by an English translation, and the rest are in English. We choose to work on the translated French portion of the data. This portion has over 8 times more documents than the English subcorpus, but with a significantly smaller portion of documents linked to an influence campaign, less than 8\%. We believe this represents a more realistic and challenging setting for detecting influence campaigns from documents. 

Given the overall small size of the dataset, we randomly split the French data into two parts, the train and test sets, but the train set can be further split for training and validation where needed.  We split at the document level, with a ratio of 80/20, as shown in Table~\ref{tab:data}. Appendix~\ref{app:data} provides further details about the distribution of media forms and average document length in the data.

\begin{table}
\centering
\begin{tabular}{ccc}
\hline
 & Train & Test\\
\hline
\textbf{\# Docs} & 5334  & 1333  \\
 & {\small (416; 7.8\%)} &  {\small (56; 4.2\%)} \\

\textbf{\# Sents} &  72,330 & 14,370  \\ 
 & {\small (15,394; 21.3\%)} & {\small (2,182; 15.2\%)} \\
 
\textbf{\# Targets$_\mathrm{ALL}$} & 270,818 & 50,781 \\ 
 & \small{(61,652; 22.8\%)} & {\small (8,531; 16.8\%)} \\
 
 \textbf{\# Targets$_\mathrm{AT}$} & 155,238 & 29,793 \\ 
 & \small{(34,703; 22.4\%)} & {\small (4,905; 16.5\%)} \\\hline
\end{tabular}
\caption{Statistics of the train and test sets (disjoint). Numbers inside  parentheses show the number and percentage of documents or document parts linked to an influence campaign. Targets$_\mathrm{AT}$: targets the \textbf{A}uthor herself believes to be \textbf{T}rue. Targets$_\mathrm{ALL}$: all belief targets, regardless of the belief holder and the commitment level.}
\label{tab:data}
\end{table}

\section{Experiments \label{sec:experiments}}

\subsection{Task description}

As argued in Sec~\ref{sec:introduction}, the real challenge for detecting influence campaigns is how to capture the phenomenon of influence campaigns. Making the detection of influence campaigns a binary classification task, i.e., to predict whether a text or a document is part of an influence campaign, is not only less realistic but also probably doomed; such a classification approach cannot survive the constantly shifting and evolving nature of influence campaigns as a dynamic social phenomenon.

Nevertheless, to comprehensively evaluate our new pipeline requires a large-scale dataset annotated on the document collection level, indicating if a collection of documents presents an influence campaign. Since the dataset described in Sec~\ref{sec:data} is the only dataset in this regard, this comprehensive evaluation cannot be possible.

Instead, we have to resort to the detection of influence campaigns as a binary classification task at the document level. This allows us to quantitatively compare our clustering approach with the existing classification approach and validate its potential as a general method to detect and characterize influence campaigns from documents. 

As a classification task, the objective is to accurately identify as many documents as possible that are linked to the known bioweapons influence campaign. Since such documents in our dataset are rare, we use precision, recall, and binary F1 to measure the classification performance of the examined approaches, which also helps us to understand the types of errors these approaches make.

\subsection{Baselines}

We train two direct document-level classifiers ({\bf Direct-document}), using fully connected feedforward neural networks (FNN) and XGBoost algorithm \citep{xgboost}, as the baselines to compare with our approach. XGBoost is an optimized gradient boosting \citep{Jerome2001} system using tree ensembles that achieve state-of-the-art results on many real-world machine learning challenges. We refrain from using any word-embedding-based or content-based machine learning models to prevent models from learning from general lexical data, which precludes the use of models such as LLMs, BERT, LSTM, and so on. We use 95 general linguistic features, extracted by an open-sourced corpus linguistics tool \citep{zwthesis}, to train the models. These features are mostly based on the work of \citet{Biber1988, Biber2006} and have been developed over decades to suit \emph{general text analysis}. We also add ``number of words'' as a feature to factor in document length, particularly for short documents (say, a tweet), which may not see the occurrence of many features at all due to length limitation. The extracted features are mostly normalized frequency counts.  More details about the model parameters and these features are given in Appendix~\ref{app:experiments}. 

In addition, we apply our pipeline on the document level ({\bf Document-level}), namely, clustering the entire documents, as an additional baseline to emphasize the importance of clustering document parts. The experimental setup for document-level clustering is identical to the setup for our approach based on document parts.

\subsection{Our approach \label{sec:ourApproach}}

\paragraph{Obtaining document parts} We break down a document into parts in three ways.  (1) We use the sentences of the document as its parts ({\bf Sentence-level}). We use the default sentence segmentation algorithm from spaCy V3.5.3. \citep{honnibal2020spacy}  (2) We also experiment with the state-of-the-art event factuality prediction system \citep{beleaf} to extract from each sentence of a document (\emph{source, target, factuality
label}) triplets. Here, \emph{source} refers to the belief holder, \emph{target} is a head word denoting an event, and \emph{factuality
label} describes the extent to which the source believes that the event has happened, is happening, or will happen. The source can either be the author herself, or somebody else according to the author.  The factuality label has five possible values, ranging from committed belief (certain that true) to committed disbelief (certain that false), with possible belief, unknown belief, and possible disbelief in between. We use a head-to-span algorithm to extract a multi-word text span, of which the target is the syntactic head, as the representation of the identified event to be used as the extracted document parts.  For {\bf Target$_\mathrm{ALL}$-level}, we use all target spans extracted by the belief system.  (3) We use the same event factuality prediction system but we retain only the events believed by the author ({\bf Target$_\mathrm{AT}$-level}), a subset of all the events identified by the event factuality prediction system, to see if document parts to which the author holds a belief will lead to a better result using our approach. The examples in Table~\ref{tab:runningExamples} are events believed by the author, where the target words are highlighted in bold. The number of sentences and targets in the train and test sets is listed in Table~\ref{tab:data}.

\paragraph{Clustering} We use S-BERT \citep{sbert} to embed document parts and then employ two clustering algorithms, i.e., KMEANS \citep{kmeans} and HDBSCAN \cite{hdbscan}, to cluster the embedded document parts. HDBSCAN is a hierarchical extension of DBSCAN \citep{dbscan} with various optimization methods implemented \cite{hdbscan}. Due to the curse of dimensionality \cite{Bellman1961}, HDBSCAN does not easily produce clusters without an embedding reduction algorithm in place. We use the state-of-the-art UMAP algorithm \citep{umap} for this purpose.

\paragraph{Classifying high-influence clusters} To contrast the baselines with our approach, we use the same two classification algorithms (FNN and XGBoost) to classify high-influence clusters. In addition to the 95 general linguistic features, there are 7 cluster-level features that are specific to our current pipeline: top-10 uni-gram text frequency, top-10 bi-gram text frequency, top-10 tri-gram text frequency, weighted n-grams text frequency, average cosine similarity between all pairs of document parts in the cluster, percentage of unique documents, and cluster size. Top-10 n-grams text frequency is the average ratio of texts containing the top-10 n-grams, whereas the weighted n-grams text frequency is the weighted sum of the aforementioned top-10 n-grams text frequencies, where the weights are simply given as $\frac{n}{\sum_{n'=1}^{3}n'}$ for $n \in \{1,2,3\}$. Average cosine similarity (ACS) is the average cosine similarity of all unique text pairs in a cluster: 

$$\mathrm{ACS} = \frac{\sum_{i}^{m} \sum_{j \neq i}^{m} \text{cos\_sim} (\text{text}_{i}, \text{text}_{j}) } {m(m-1)}$$

\noindent These 5 features are designed as ``hard'' (ngrams) and ``soft'' (ACS) measurements of topical and thematic coherence of a cluster, which are relational and independent of the specific lexical choices used inside the cluster. Percentage of unique documents and cluster size are just basic attributes of a cluster. The percentage of unique documents is calculated by dividing the number of  documents whose parts occur in the cluster by the number of document parts in the cluster.

\paragraph{Cluster aggregation} We run a total of 135 clustering experiments by varying related parameters for the two clustering algorithms we use, detailed in the Appendix~\ref{app:experiments}. We use all  resulting clusters both for training the high-influence cluster classifier, as well as for selecting high-influence documents from high-influence clusters. 

\paragraph{Classifying high-influence documents} For a single clustering experiment, we simply classify any document, whose parts occur in at least one high-influence cluster, as a high-influence document. This low threshold is to ensure that short documents would not be excluded from being identified as high-influence documents, since they may only be segmented into one part and cannot have more than one association with high-influence clusters. 

As mentioned, we propose using all high-influence clusters from multiple clustering experiments to expand the search for high-influence documents. However, as a result of this aggregation, the chance of misidentifying a high-influence document based on a single association with a high-influence cluster increases, since false positives in high-influence clusters also accumulate with aggregation. To regulate the false positive rate, we set $\beta=0.2$ (see Sec~\ref{sec:pipeline} for definition).

\paragraph{Evaluation} Since a clustering configuration does not necessarily produce high-influence clusters and in practice we can always try to find one that does, we evaluate the average performance of our approach on various clustering setups that produce high-influence clusters predicted by our pipeline. We choose a wide range of clustering configurations (see Appendix~\ref{app:experiments}) so as to avoid hard finetuning our approach. We run the two classification algorithms (FNN and XGBoost) for five times with varying parameters to compare the average performance of our approach against the baseline approaches on our dataset.

\section{Results \label{sec:results}}

\begin{table*}
\centering
\begin{tabular}{llllllll}
 & \multicolumn{3}{c}{FNN} & & \multicolumn{3}{c}{XGBoost} \\
\hline
& \textbf{Precision} & \textbf{Recall} & \textbf{F1} & & \textbf{Precision} & \textbf{Recall} & \textbf{F1}\\
\hline
Direct-document & 20.2$_{\pm2.2}$  & 18.9$_{\pm14.9}$ & 17.1$_{\pm8.6}$ & & 77.3$_{\pm9.3}$  & 37.9$_{\pm7.9}$ & 50.7$_{\pm9.1}$ \\ \hline\hline

Document-level (mean) & 0.3$_{\pm0.1}$  & 0.7$_{\pm0.1}$ & 0.4$_{\pm0.1}$ & & 90.7$_{\pm2.5}$  & 25.4$_{\pm3.3}$ & 38.2$_{\pm4.6}$ \\
{\small + Aggregation} & 0.0$_{\pm0.0}$  & 0.0$_{\pm0.0}$ & 0.0$_{\pm0.0}$ & & 94.1$_{\pm0.7}$  & 28.6$_{\pm3.1}$ & 43.8$_{\pm3.8}$ \\\hline

Sentence-level (mean) & 28.3$_{\pm4.1}$  & 44.1$_{\pm4.7}$ & 32.8$_{\pm4.2}$ & & 69.4$_{\pm10.9}$  & 50.4$_{\pm2.7}$ & 56.7$_{\pm4.1}$ \\
{\small + Aggregation} & 74.5$_{\pm16.4}$  & 43.2$_{\pm4.1}$ & 54.3$_{\pm7.7}$ & & 86.5$_{\pm1.8}$  & 70.7$_{\pm2.4}$ & 77.8$_{\pm2.0}$ \\\hline

Target$_\mathrm{ALL}$-level (mean) & 25.4$_{\pm6.7}$  & 35.2$_{\pm8.5}$ & 27.0$_{\pm6.9}$ & & 78.2$_{\pm3.7}$  & 73.8$_{\pm2.4}$ & 75.3$_{\pm1.3}$ \\
{\small + Aggregation}  & 72.5$_{\pm4.5}$  & 40.0$_{\pm5.7}$ & 51.5$_{\pm5.7}$ & & 81.1$_{\pm3.5}$  & 71.1$_{\pm7.3}$ & 75.5$_{\pm3.8}$ \\\hline

Target$_\mathrm{AT}$-level (mean) & 60.7$_{\pm7.1}$  & 66.8$_{\pm10.5}$ & 62.4$_{\pm8.5}$ &  & 63.5$_{\pm2.2}$  & 49.5$_{\pm2.8}$ & 54.8$_{\pm2.1}$ \\
{\small + Aggregation}  & 64.8$_{\pm4.6}$  & 61.8$_{\pm8.6}$ & 63.1$_{\pm6.0}$ & & 80.2$_{\pm3.5}$  & 71.4$_{\pm1.8}$ & 75.5$_{\pm0.9}$  \\\hline
\end{tabular}
\caption{Average model test set performance plus standard deviation (in \%) under different training conditions for five runs. For the clustering approach, when there is no aggregation of high-influence clusters, the performance is the average performance across different clustering experiments averaged over five runs. Note that the two classification algorithms (FNN and XGBoost) are only used to classify clusters in our approach.}
\label{tab:mainResults}
\end{table*}

\subsection{Main findings}

\paragraph{The document classification approach versus ours} Table~\ref{tab:mainResults} shows the main results of the experiments. As expected, our approach significantly outperforms the direct document-level classification approach.

\paragraph{Clustering documents versus document parts} Clustering document parts clearly outperforms clustering documents by a significant margin. When FNN is used to classify high-influence clusters, clustering documents barely works at all.

\paragraph{FNN versus XGBoost} Our models that use XGBoost to classify clusters achieve overall high precision, regardless of aggregation. Those which use FNN suffer from low precision without aggregation. This means that high-influence clusters predicted by XGBoost contain much fewer false positives, i.e., associated documents with no link to an influence campaign, than those predicted by FNN. This makes XGBoost a better choice for cluster prediction for the current paper.

\paragraph{Document parts} There is value in clustering belief targets. They are multi-word text spans within a sentence that carry a factuality label and involve a belief source. Compared to full sentences, they are more information-dense. We find that when FNN is used to classify clusters, clustering belief targets the author holds to be true (Target$_\mathrm{AT}$-level) leads to the best performance, independent of the use of aggregation. When XGBoost is used, clustering all belief targets outperforms clustering sentences by 18\% absolute in F1 without aggregation. These results show the potential of extracting belief targets for a better detection of influence campaigns, which intuitively make sense because influence campaigns are all about spreading a certain belief of the influencers. 

\paragraph{Cluster aggregation} 
From Table~\ref{tab:mainResults}, we see that cluster aggregation helps in every experiment (leaving aside document-level clustering using FNN, which performs at near-0 levels).  Most of these improvements are statistically significant, given the standard deviations shown.  However, for FNN using Target$_\mathrm{AT}$-level there is no significant difference, and for XGBoost using Target$_\mathrm{ALL}$-level there is no significant difference.  We have no explanation for these exceptions for now.  In general, cluster aggregation helps in two ways. When our models have very low precision (using FNN), the current aggregation setup rules out many false positives resulting from misclassified high-influence clusters, which greatly improves precision. On the other hand, when precision is decent (in the case of XGBoost), aggregation can serve to help increase the range of relevant documents associated with high-influence clusters, which lead to a better recall in most cases.

\subsection{Error analysis}

\begin{table}
\centering
\small
\setlength{\tabcolsep}{1pt}
\begin{tabular}{l|rr|rr|rr}
\hline
Media (pos/neg) & \multicolumn{2}{c|}{Direct-Doc} & \multicolumn{2}{c|}{Doc-level} & \multicolumn{2}{c}{Our approach} \\
 & FN & FP & FN & FP & FN & FP \\
\hline
\textbf{Twitter} (11/686) & 11.0 & 0 & 8.8 & 0 & 3.0  & 1.0  \\
\textbf{Forum} (7/136) & 5.4  & 0 & 6.0 & 0 
 & 1.0  & 0  \\
\textbf{News} (24/280) & 11.4  & 4.6 & 17.0 & 0 & 10.4  & 4.2 \\
\textbf{Blog} (13/62) & 6.0 & 1.4 & 7.8 & 1.0 & 2.0  & 1.0  \\
\textbf{Reddit} (0/91) & NA & 0 & NA & 0 & NA  & 0 \\
\textbf{Other} (1/22) & 1.0 & 0 & 0.4 & 0 & 0  & 0  \\
\textbf{Total} (56/1277) & 34.8 & 6.0 & 40.0 & 1.0 & 16.4  & 6.2  \\\hline
\end{tabular}
\caption{Average test set error counts of the three best models from the two baseline approaches plus ours across the six media types over the five runs. The first column indicate the media type along with the numbers of documents that reflect (``pos'') or do not reflect (``neg'') an influence campaign. FN: False Negative. FP: False Positive. NA is due to zero positives in Reddit. The best model configurations for the three approaches all use XGBoost and aggregation where applicable. Our best system is on the sentence level per Table~\ref{tab:mainResults}.}
\label{tab:errorCounts}
\end{table}

Table~\ref{tab:errorCounts} reports the average number of errors made by the two baseline approaches and our approach on the test set, over the five runs. The error counts are broken down according to media types.  

\paragraph{Direct-document} The direct document-level classification approach fails to recognize all the documents from Twitter and Other that reflect the bioweapons influence campaign. It also misses 5.4 out of 7 documents from Forum on average across five runs, with slightly less than 50\% false negative rate for documents from News and Blog. Given the average document length for these five media types (see Table~\ref{tab:metadata}), it is clear that this classification approach works poorly on identifying short documents linked to an influence campaign when training on documents with a wide length range. This is probably because the model learns some discriminative features from long documents, which may not be observed in short documents. Conversely, a similar issue may also occur the other way around, suppose the model trains on short documents. This may be one of the inherent limitations of the direct document-level classification approach, even when the models are trained and deployed for predicting an known influence campaign. 

\paragraph{Document-level} Directly clustering documents allows the model to recognize influence campaigns in documents of different genres and length, which is an advantage compared to the direct document-level classification approach. However, clustering documents also makes it hard for the model to efficiently identify information in the documents related to influence campaigns, which may exist only in parts of document. This results in the high number of false negatives. That said, our current pipeline setup helps this document-level clustering approach to make accurate positive predictions, given the lowest number of false positives.

\paragraph{Our approach} Clearly, clustering document parts helps overcome the limitations faced by the two baseline approaches, since the model recognizes influence campaigns in documents irrespective of the genre and have less than half false negatives, compared to the other two approaches. We identify 19 documents (15 FNs and 4 FPs) where the models across the five runs consistently misclassify. By using the keyword ``bio'', we identify 3 of the 19 documents that may be mislabelled. For the remaining 16 misclassifed documents, we hypothesize that the errors are mainly caused by two reasons. First, sentences do not necessarily reflect the theme of the document, which, for example, may make our model confuse documents exposing an influence campaign with one that spreads it. Second, none of the other techniques (e.g., SBERT, clustering) used in our pipeline are free of errors, which can propagate and ultimately lead to a wrong classification decision. We wish to improve our pipeline along these two directions in the future.

\begin{figure}
    \centering
\includegraphics[width=1\columnwidth]{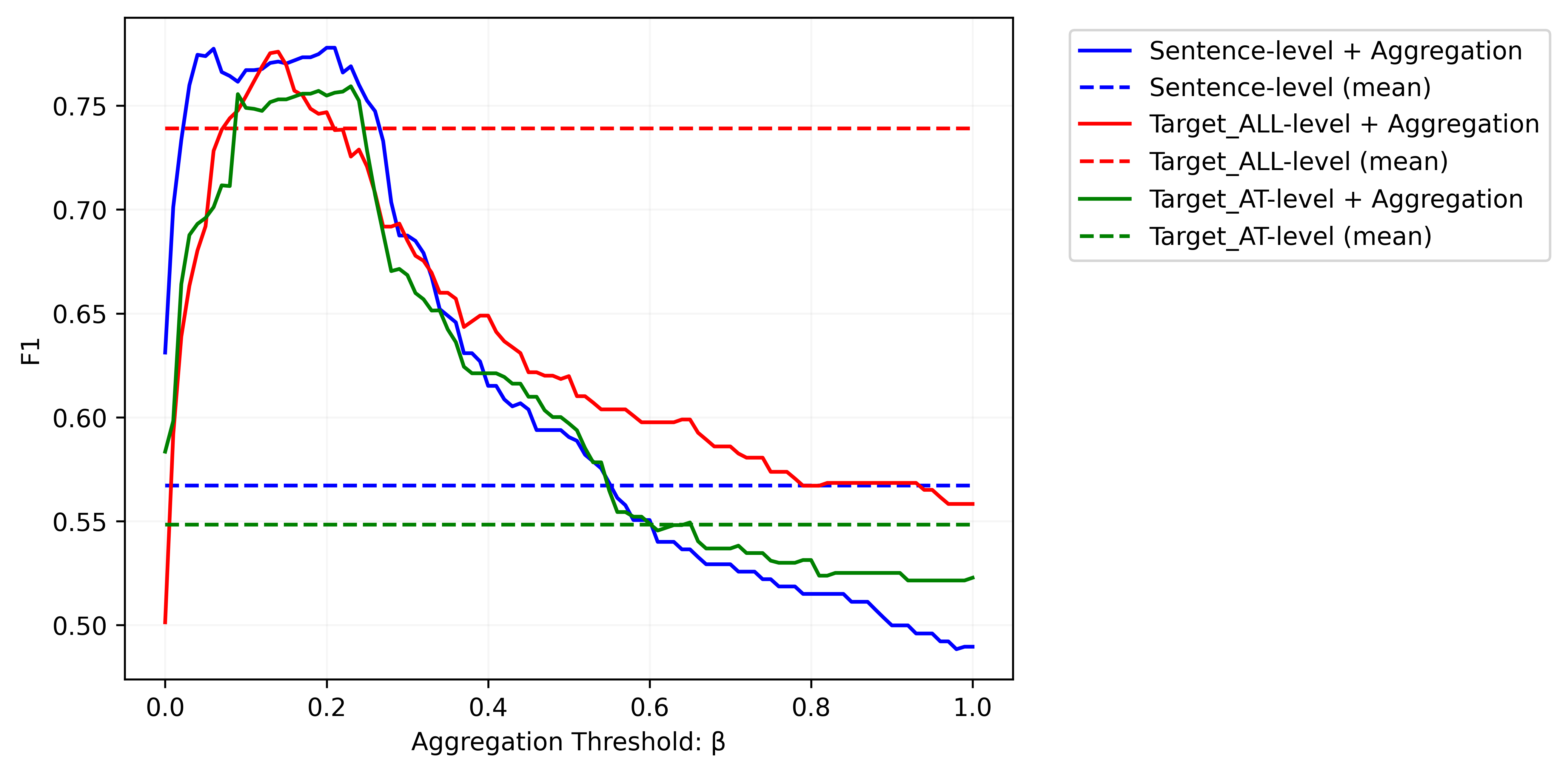}
    \caption{Aggregation versus no Aggregation with XGBoost as the high-influence cluster classifier. Results are averaged over the five runs.}
    \label{fig:aggregationXGBoost}
\end{figure}

\subsection{Threshold for models with aggregation}

Concerns may arise over our use of a less justified threshold $\beta$ to select documents from high-influence clusters as high-influence documents. This threshold is is a ratio of the number of high-influence clusters available for aggregation. As explained in Sec~\ref{sec:pipeline}, this ratio helps prevent the false negatives in each high-influence cluster from accumulating uncontrolled, as a result of aggregation.   

Nevertheless, as shown in Fig~\ref{fig:aggregationXGBoost} (also see Fig~\ref{fig:aggregationFNN} from Appendix~\ref{app:results} for FNN), the classification F1 with aggregation is almost always better than without aggregation for most models. Unsurprisingly, the performance curve shows an upside down U-shape, a trade-off between precision and recall as we vary $\beta$. Setting $\beta = 0.2$ is a conservative choice, which does not lead to the optimal performance. In the future, we would like to explore an automatic way of finding the optimal value for $\beta$.







%





\section{Conclusion \label{sec:discussion}}

We have presented a new approach to finding influence campaigns, which relies on four core features: (1) we cluster parts of documents; (2) we classify clusters of parts of documents using non-lexical features; (3) we relate the classification result back to documents; (4) we use cluster aggregation, the use of many clustering runs over the same dataset, to augment training data for the cluster classifier. The resulting classification of the documents does not only show a predicted label for the document (part of influence campaign or not), but it also shows which parts of the document are responsible for this classification. We believe that our general approach can profit other document classification tasks, including detecting scientific influence in published papers, or themes in literature.

There are several avenues for possible future work and we list three below. (1) \textbf{Datasets}. Given the increasing importance of detecting influence campaigns, we hope there will be more datasets annotated on the document collection level for an influence campaign. (2) \textbf{Incorporating non-textual information}. Our current pipeline is a text-only system. Leveraging non-textual information, such as social interactions and the authors' past activities, may help us create a more complicated and comprehensive system (e.g., using graph neural network) that enhances the accurate and reliable detection of influence campaigns. However, such work cannot be possible without good datasets. (3) \textbf{Automatic characterization of influence campaigns}. Our work captures influence campaigns by the high-influence clusters, which may contain a large number of semantically related document parts, possibly with noise. To fully make sense of these clusters, we need to have some automatic ways of characterizing them in a fine-grained and interpretable way aligned with the downstream needs. Our preliminary experiments show that LLMs may be a potential option.

\section*{Acknowledgement}

We thank three anonymous reviewers from the 6th NLP+CSS Workshop for the constructive and helpful comments. This material is based on work supported by the Defense Advanced Research Projects Agency (DARPA) under Contracts No. HR01121C0186, No. HR001120C0037, and PR No. HR0011154158. Any opinions, findings
and conclusions or recommendations expressed in this material are those of the authors and do not
necessarily reflect the views of DARPA. Rambow gratefully acknowledges support from the Institute for Advanced Computational Science at Stony Brook University.

\section*{Limitations}

This study serves as a preliminary evaluation and validation of the new paradigm we propose for influence campaign detection. Given the lack of data and constrained by time, we have not been able to show that the approach also works on an entirely unseen dataset (though of course we tested it on unseen documents in our dataset). 

\begin{itemize}
    \item We define influence campaigns in a very general sense, but our approach is only tested on data relating to political influence campaigns. We need to test our approach on other non-political influence campaign datasets. 
    \item We cannot release the dataset we used to train and test our pipeline due to the funding agency's restrictions. We hope once the current program is finalized, the dataset will be released so that our study can be reproduced.

    \item We did not spend a large amount of time attempting to improve the direct-document approach.  We cannot guarantee that with a different set of (non-lexical) features and well-tuned parameters, a direct document-level classifier may not outperform our approach. 
\end{itemize}

\section*{Ethical Concerns}

Working with social media often brings privacy concerns.  The data we are working with has already been anonymized.  For example, Twitter handles have been replaced by random designators.  Furthermore, in our work, we do not use any part of the information about the author, we only use the text.

\bibliography{references}

\begin{thebibliography}{31}
\expandafter\ifx\csname natexlab\endcsname\relax\def\natexlab#1{#1}\fi

\bibitem[{Addawood et~al.(2019)Addawood, Badawy, Lerman, and
  Ferrara}]{Addawood2019LinguisticCT}
Aseel Addawood, Adam Badawy, Kristina Lerman, and Emilio Ferrara. 2019.
\newblock Linguistic cues to deception: Identifying political trolls on social
  media.
\newblock In \emph{International Conference on Web and Social Media}.

\bibitem[{Badawy et~al.(2018)Badawy, Ferrara, and Lerman}]{Badawy2018}
Adam Badawy, Emilio Ferrara, and Kristina Lerman. 2018.
\newblock \href {https://doi.org/10.1109/ASONAM.2018.8508646} {Analyzing the
  digital traces of political manipulation: The 2016 russian interference
  twitter campaign}.
\newblock In \emph{2018 IEEE/ACM International Conference on Advances in Social
  Networks Analysis and Mining (ASONAM)}, pages 258--265.

\bibitem[{Barrón-Cedeño et~al.(2019)Barrón-Cedeño, Jaradat, {Da San
  Martino}, and Nakov}]{Proppy2019}
Alberto Barrón-Cedeño, Israa Jaradat, Giovanni {Da San Martino}, and Preslav
  Nakov. 2019.
\newblock \href {https://doi.org/https://doi.org/10.1016/j.ipm.2019.03.005}
  {Proppy: Organizing the news based on their propagandistic content}.
\newblock \emph{Information Processing \& Management}, 56(5):1849--1864.

\bibitem[{Bellman(1961)}]{Bellman1961}
Richard~E. Bellman. 1961.
\newblock \emph{Adaptive control processes: a guided tour}.
\newblock Princeton University Press.

\bibitem[{Biber(1988)}]{Biber1988}
Douglas Biber. 1988.
\newblock \emph{Variation across Speech and Writing}.
\newblock Cambridge University Press.

\bibitem[{Biber(2006)}]{Biber2006}
Douglas Biber. 2006.
\newblock \emph{University Language: A Corpus-Based Study of Spoken and Written
  Registers}.
\newblock John Benjamins Publishing.

\bibitem[{Campello et~al.(2015)Campello, Moulavi, Zimek, and Sander}]{hdbscan}
Ricardo J. G.~B. Campello, Davoud Moulavi, Arthur Zimek, and J\"{o}rg Sander.
  2015.
\newblock \href {https://doi.org/10.1145/2733381} {Hierarchical density
  estimates for data clustering, visualization, and outlier detection}.
\newblock 10(1).

\bibitem[{Chen and Guestrin(2016)}]{xgboost}
Tianqi Chen and Carlos Guestrin. 2016.
\newblock \href {https://doi.org/10.1145/2939672.2939785} {Xgboost: A scalable
  tree boosting system}.
\newblock In \emph{Proceedings of the 22nd ACM SIGKDD International Conference
  on Knowledge Discovery and Data Mining}, KDD '16, page 785–794, New York,
  NY, USA. Association for Computing Machinery.

\bibitem[{Davis et~al.(2016)Davis, Varol, Ferrara, Flammini, and
  Menczer}]{Davis2016}
Clayton~Allen Davis, Onur Varol, Emilio Ferrara, Alessandro Flammini, and
  Filippo Menczer. 2016.
\newblock \href {https://doi.org/10.1145/2872518.2889302} {Botornot: A system
  to evaluate social bots}.
\newblock In \emph{Proceedings of the 25th International Conference Companion
  on World Wide Web}, WWW '16 Companion, page 273–274, Republic and Canton of
  Geneva, CHE. International World Wide Web Conferences Steering Committee.

\bibitem[{des Mesnards and Zaman(2018)}]{mesnards2018detecting}
Nicolas~Guenon des Mesnards and Tauhid Zaman. 2018.
\newblock \href {http://arxiv.org/abs/1805.10244} {Detecting influence
  campaigns in social networks using the ising model}.

\bibitem[{Ester et~al.(1996)Ester, Kriegel, Sander, and Xu}]{dbscan}
Martin Ester, Hans-Peter Kriegel, J\"{o}rg Sander, and Xiaowei Xu. 1996.
\newblock A density-based algorithm for discovering clusters in large spatial
  databases with noise.
\newblock In \emph{Proceedings of the Second International Conference on
  Knowledge Discovery and Data Mining}, KDD'96, page 226–231. AAAI Press.

\bibitem[{Ferrara(2017)}]{Ferrara_2017}
Emilio Ferrara. 2017.
\newblock \href {https://doi.org/10.5210/fm.v22i8.8005} {Disinformation and
  social bot operations in the run up to the 2017 french presidential
  election}.
\newblock \emph{First Monday}.

\bibitem[{Friedman(2001)}]{Jerome2001}
Jerome~H. Friedman. 2001.
\newblock \href {https://doi.org/10.1214/aos/1013203451} {{Greedy function
  approximation: A gradient boosting machine.}}
\newblock \emph{The Annals of Statistics}, 29(5):1189 -- 1232.

\bibitem[{Hajli et~al.(2022)Hajli, Saeed, Tajvidi, and Shirazi}]{Hajli2022}
Nick Hajli, Usman Saeed, Mina Tajvidi, and Farid Shirazi. 2022.
\newblock \href {https://doi.org/https://doi.org/10.1111/1467-8551.12554}
  {Social bots and the spread of disinformation in social media: The challenges
  of artificial intelligence}.
\newblock \emph{British Journal of Management}, 33(3):1238--1253.

\bibitem[{Himelein-Wachowiak et~al.(2021)Himelein-Wachowiak, Giorgi, Devoto,
  Rahman, Ungar, Schwartz, Epstein, Leggio, and Curtis}]{Himelein2021}
McKenzie Himelein-Wachowiak, Salvatore Giorgi, Amanda Devoto, Muhammad Rahman,
  Lyle Ungar, H.~Schwartz, David Epstein, Lorenzo Leggio, and Brenda Curtis.
  2021.
\newblock \href {https://doi.org/10.2196/26933} {Bots and misinformation spread
  on social media: A mixed scoping review with implications for covid-19
  (preprint)}.
\newblock \emph{Journal of Medical Internet Research}, 23.

\bibitem[{Honnibal et~al.(2020)Honnibal, Montani, Van~Landeghem, and
  Boyd}]{honnibal2020spacy}
Matthew Honnibal, Ines Montani, Sofie Van~Landeghem, and Adriane Boyd. 2020.
\newblock \href {https://doi.org/10.5281/zenodo.1212303} {{spaCy:
  Industrial-strength Natural Language Processing in Python}}.

\bibitem[{Littman et~al.(2016)Littman, Wrubel, and Kerchner}]{PDI7IN2016}
Justin Littman, Laura Wrubel, and Daniel Kerchner. 2016.
\newblock \href {https://doi.org/10.7910/DVN/PDI7IN} {{2016 United States
  Presidential Election Tweet Ids}}.

\bibitem[{Luceri et~al.(2023)Luceri, Boniardi, and
  Ferrara}]{luceri2023leveraging}
Luca Luceri, Eric Boniardi, and Emilio Ferrara. 2023.
\newblock \href {http://arxiv.org/abs/2311.07816} {Leveraging large language
  models to detect influence campaigns in social media}.

\bibitem[{MacQueen(1967)}]{kmeans}
J.~MacQueen. 1967.
\newblock \href {https://api.semanticscholar.org/CorpusID:6278891} {Some
  methods for classification and analysis of multivariate observations}.

\bibitem[{Malik et~al.(2023)Malik, Imran, and Mamdouh}]{Malik2023}
Muhammad Shahid~Iqbal Malik, Tahir Imran, and Jamjoom~Mona Mamdouh. 2023.
\newblock \href {https://doi.org/10.7717/peerj-cs.1248} {How to detect
  propaganda from social media? exploitation of semantic and fine-tuned
  language models}.
\newblock \emph{PeerJ Comput Sci}.

\bibitem[{Martin et~al.(2023)Martin, Shapiro, and Ilhardt}]{Diego2023}
Diego~A Martin, Jacob~N Shapiro, and Julia~G Ilhardt. 2023.
\newblock \href {https://doi.org/10.1177/00223433221092815} {Introducing the
  online political influence efforts dataset}.
\newblock \emph{Journal of Peace Research}, 60(5):868--876.

\bibitem[{McInnes et~al.(2018)McInnes, Healy, Saul, and Großberger}]{umap}
Leland McInnes, John Healy, Nathaniel Saul, and Lukas Großberger. 2018.
\newblock \href {https://doi.org/10.21105/joss.00861} {Umap: Uniform manifold
  approximation and projection}.
\newblock \emph{Journal of Open Source Software}, 3(29):861.

\bibitem[{Mekontchou et~al.(2023)Mekontchou, Fotsoh, Batchakui, and
  Ella}]{mekontchou2023information}
Paul~Mbate Mekontchou, Armel Fotsoh, Bernabe Batchakui, and Eddy Ella. 2023.
\newblock \href {http://arxiv.org/abs/2302.10150} {Information retrieval in
  long documents: Word clustering approach for improving semantics}.

\bibitem[{Murzaku et~al.(2023)Murzaku, Osborne, Aviram, and Rambow}]{beleaf}
John Murzaku, Tyler Osborne, Amittai Aviram, and Owen Rambow. 2023.
\newblock \href {https://doi.org/10.18653/v1/2023.findings-acl.44} {Towards
  generative event factuality prediction}.
\newblock In \emph{Findings of the Association for Computational Linguistics:
  ACL 2023}, pages 701--715, Toronto, Canada. Association for Computational
  Linguistics.

\bibitem[{Nogara et~al.(2022)Nogara, Vishnuprasad, Cardoso, Ayoub, Giordano,
  and Luceri}]{Nogara2022}
Gianluca Nogara, Padinjaredath~Suresh Vishnuprasad, Felipe Cardoso, Omran
  Ayoub, Silvia Giordano, and Luca Luceri. 2022.
\newblock \href {https://doi.org/10.1145/3501247.3531573} {The disinformation
  dozen: An exploratory analysis of covid-19 disinformation proliferation on
  twitter}.
\newblock In \emph{Proceedings of the 14th ACM Web Science Conference 2022},
  WebSci '22, page 348–358, New York, NY, USA. Association for Computing
  Machinery.

\bibitem[{Reimers and Gurevych(2019)}]{sbert}
Nils Reimers and Iryna Gurevych. 2019.
\newblock \href {https://doi.org/10.18653/v1/D19-1410} {Sentence-{BERT}:
  Sentence embeddings using {S}iamese {BERT}-networks}.
\newblock In \emph{Proceedings of the 2019 Conference on Empirical Methods in
  Natural Language Processing and the 9th International Joint Conference on
  Natural Language Processing (EMNLP-IJCNLP)}, pages 3982--3992, Hong Kong,
  China. Association for Computational Linguistics.

\bibitem[{Rossetti and Zaman(2022)}]{Rossetti2022BotsDA}
Michael Rossetti and Tauhid Zaman. 2022.
\newblock \href {https://api.semanticscholar.org/CorpusID:253099441} {Bots,
  disinformation, and the first impeachment of u.s. president donald trump}.
\newblock \emph{PLOS ONE}, 18.

\bibitem[{Rubin(2017)}]{Rubin2017DeceptionDA}
Victoria~L. Rubin. 2017.
\newblock \href {https://api.semanticscholar.org/CorpusID:151727586} {Deception
  detection and rumor debunking for social media}.

\bibitem[{Sakketou et~al.(2022)Sakketou, Plepi, Cervero, Geiss, Rosso, and
  Flek}]{sakketou-etal-2022-factoid}
Flora Sakketou, Joan Plepi, Riccardo Cervero, Henri~Jacques Geiss, Paolo Rosso,
  and Lucie Flek. 2022.
\newblock \href {https://aclanthology.org/2022.lrec-1.345} {{FACTOID}: A new
  dataset for identifying misinformation spreaders and political bias}.
\newblock In \emph{Proceedings of the Thirteenth Language Resources and
  Evaluation Conference}, pages 3231--3241, Marseille, France. European
  Language Resources Association.

\bibitem[{Wang(2021)}]{zwthesis}
Zhengxiang Wang. 2021.
\newblock A macroscopic re-examination of language and gender: A corpus-based
  case study in university instructor discourses.
\newblock Master's thesis, University of Saskatchewan.

\bibitem[{Álvaro Figueira and Oliveira(2017)}]{FIGUEIRA2017817}
Álvaro Figueira and Luciana Oliveira. 2017.
\newblock \href {https://doi.org/https://doi.org/10.1016/j.procs.2017.11.106}
  {The current state of fake news: challenges and opportunities}.
\newblock \emph{Procedia Computer Science}, 121:817--825.
\newblock CENTERIS 2017 - International Conference on ENTERprise Information
  Systems / ProjMAN 2017 - International Conference on Project MANagement /
  HCist 2017 - International Conference on Health and Social Care Information
  Systems and Technologies, CENTERIS/ProjMAN/HCist 2017.

\end{thebibliography}


\begin{thebibliography}{17}
\expandafter\ifx\csname natexlab\endcsname\relax\def\natexlab#1{#1}\fi

\bibitem[{Bhagat and Hovy(2013)}]{bhagat-hovy-2013-squibs}
Rahul Bhagat and Eduard Hovy. 2013.
\newblock \href {https://doi.org/10.1162/COLI_a_00166} {{S}quibs: What is a
  paraphrase?}
\newblock \emph{Computational Linguistics}, 39(3):463--472.

\bibitem[{Feng et~al.(2021)Feng, Gangal, Wei, Chandar, Vosoughi, Mitamura, and
  Hovy}]{feng-etal-2021-survey}
Steven~Y. Feng, Varun Gangal, Jason Wei, Sarath Chandar, Soroush Vosoughi,
  Teruko Mitamura, and Eduard Hovy. 2021.
\newblock \href {https://doi.org/10.18653/v1/2021.findings-acl.84} {A survey of
  data augmentation approaches for {NLP}}.
\newblock In \emph{Findings of the Association for Computational Linguistics:
  ACL-IJCNLP 2021}, pages 968--988, Online. Association for Computational
  Linguistics.

\bibitem[{Hou et~al.(2018)Hou, Liu, Che, and Liu}]{hou2018}
Yutai Hou, Yijia Liu, Wanxiang Che, and Ting Liu. 2018.
\newblock \href {http://arxiv.org/abs/1807.01554} {Sequence-to-sequence data
  augmentation for dialogue language understanding}.

\bibitem[{Iwana and Uchida(2021)}]{Iwana2021}
Brian~Kenji Iwana and Seiichi Uchida. 2021.
\newblock \href {https://doi.org/10.1371/journal.pone.0254841} {An empirical
  survey of data augmentation for time series classification with neural
  networks}.
\newblock \emph{PLOS ONE}, 16(7):1--32.

\bibitem[{Kobayashi(2018)}]{kobayashi-2018-contextual}
Sosuke Kobayashi. 2018.
\newblock \href {https://doi.org/10.18653/v1/N18-2072} {Contextual
  augmentation: Data augmentation by words with paradigmatic relations}.
\newblock In \emph{Proceedings of the 2018 Conference of the North {A}merican
  Chapter of the Association for Computational Linguistics: Human Language
  Technologies, Volume 2 (Short Papers)}, pages 452--457, New Orleans,
  Louisiana. Association for Computational Linguistics.

\bibitem[{Kurata et~al.(2016)Kurata, Xiang, and Zhou}]{Kurata2016LabeledDG}
Gakuto Kurata, Bing Xiang, and Bowen Zhou. 2016.
\newblock Labeled data generation with encoder-decoder lstm for semantic slot
  filling.
\newblock In \emph{INTERSPEECH}.

\bibitem[{Liu(2012)}]{Bing2012}
Bing Liu. 2012.
\newblock \emph{Sentiment Analysis and Opinion Mining}.
\newblock Morgan \& Claypool.

\bibitem[{Liu et~al.(2020)Liu, Wang, Xiang, and Meng}]{Liu:9240734}
Pei Liu, Xuemin Wang, Chao Xiang, and Weiye Meng. 2020.
\newblock \href {https://doi.org/10.1109/CCNS50731.2020.00049} {A survey of
  text data augmentation}.
\newblock In \emph{2020 International Conference on Computer Communication and
  Network Security (CCNS)}, pages 191--195.

\bibitem[{Liu et~al.(2018)Liu, Chen, Deng, Zeng, Chen, Li, and
  Tang}]{liu-etal-2018-lcqmc}
Xin Liu, Qingcai Chen, Chong Deng, Huajun Zeng, Jing Chen, Dongfang Li, and
  Buzhou Tang. 2018.
\newblock \href {https://aclanthology.org/C18-1166} {{LCQMC}:a large-scale
  {C}hinese question matching corpus}.
\newblock In \emph{Proceedings of the 27th International Conference on
  Computational Linguistics}, pages 1952--1962, Santa Fe, New Mexico, USA.
  Association for Computational Linguistics.

\bibitem[{Park et~al.(2019)Park, Chan, Zhang, Chiu, Zoph, Cubuk, and
  Le}]{Park2019}
Daniel~S. Park, William Chan, Yu~Zhang, Chung-Cheng Chiu, Barret Zoph, Ekin~D.
  Cubuk, and Quoc~V. Le. 2019.
\newblock \href {https://doi.org/10.21437/interspeech.2019-2680} {Specaugment:
  A simple data augmentation method for automatic speech recognition}.
\newblock \emph{Interspeech 2019}.

\bibitem[{Shorten and Khoshgoftaar(2019)}]{Shorten2019ASO}
Connor Shorten and Taghi~M. Khoshgoftaar. 2019.
\newblock A survey on image data augmentation for deep learning.
\newblock \emph{Journal of Big Data}, 6:1--48.

\bibitem[{Wang and Yang(2015)}]{wang-yang-2015-thats}
William~Yang Wang and Diyi Yang. 2015.
\newblock \href {https://doi.org/10.18653/v1/D15-1306} {That{'}s so
  annoying!!!: A lexical and frame-semantic embedding based data augmentation
  approach to automatic categorization of annoying behaviors using {\#}petpeeve
  tweets}.
\newblock In \emph{Proceedings of the 2015 Conference on Empirical Methods in
  Natural Language Processing}, pages 2557--2563, Lisbon, Portugal. Association
  for Computational Linguistics.

\bibitem[{Wang et~al.(2018)Wang, Pham, Dai, and
  Neubig}]{wang-etal-2018-switchout}
Xinyi Wang, Hieu Pham, Zihang Dai, and Graham Neubig. 2018.
\newblock \href {https://doi.org/10.18653/v1/D18-1100} {{S}witch{O}ut: an
  efficient data augmentation algorithm for neural machine translation}.
\newblock In \emph{Proceedings of the 2018 Conference on Empirical Methods in
  Natural Language Processing}, pages 856--861, Brussels, Belgium. Association
  for Computational Linguistics.

\bibitem[{Wei et~al.(2021)Wei, Huang, Xu, and Vosoughi}]{wei-etal-2021-text}
Jason Wei, Chengyu Huang, Shiqi Xu, and Soroush Vosoughi. 2021.
\newblock \href {https://doi.org/10.18653/v1/2021.eacl-main.252} {Text
  augmentation in a multi-task view}.
\newblock In \emph{Proceedings of the 16th Conference of the European Chapter
  of the Association for Computational Linguistics: Main Volume}, pages
  2888--2894, Online. Association for Computational Linguistics.

\bibitem[{Wei and Zou(2019)}]{wei-zou-2019-eda}
Jason Wei and Kai Zou. 2019.
\newblock \href {https://doi.org/10.18653/v1/D19-1670} {{EDA}: Easy data
  augmentation techniques for boosting performance on text classification
  tasks}.
\newblock In \emph{Proceedings of the 2019 Conference on Empirical Methods in
  Natural Language Processing and the 9th International Joint Conference on
  Natural Language Processing (EMNLP-IJCNLP)}, pages 6382--6388, Hong Kong,
  China. Association for Computational Linguistics.

\bibitem[{Xie et~al.(2020)Xie, Dai, Hovy, Luong, and Le}]{xie2020unsupervised}
Qizhe Xie, Zihang Dai, Eduard Hovy, Minh-Thang Luong, and Quoc~V. Le. 2020.
\newblock \href {http://arxiv.org/abs/1904.12848} {Unsupervised data
  augmentation for consistency training}.

\bibitem[{Zhang et~al.(2015)Zhang, Zhao, and LeCun}]{Zhang2015}
Xiang Zhang, Junbo Zhao, and Yann LeCun. 2015.
\newblock \href
  {https://proceedings.neurips.cc/paper/2015/file/250cf8b51c773f3f8dc8b4be867a9a02-Paper.pdf}
  {Character-level convolutional networks for text classification}.
\newblock In \emph{Advances in Neural Information Processing Systems},
  volume~28. Curran Associates, Inc.

\end{thebibliography}
\bibliographystyle{acl_natbib}

\appendix

\section{Data}
\label{app:data}

\begin{table}
\centering
\begin{tabular}{lll}
\hline
Media & \# docs {\small (train + test)} & Avg doc length\\
\hline
\textbf{Twitter} & 2,795 {\small (2,098 + 697)}  & 26.6$_{\pm 13.1}$  \\
\textbf{Forum} & 887 {\small (744 + 143)} & 330.7$_{\pm 574.4}$  \\
\textbf{News} & 2,039 {\small (1,735 + 304)}  & 654.8$_{\pm 851.9}$  \\
\textbf{Blog} & 415 {\small (340 + 75)} & 945.0$_{\pm 1,668.1}$  \\
\textbf{Reddit} & 371 {\small (280 + 91)}  & 69.3$_{\pm 131.7}$  \\
\textbf{Other} & 160 {\small (137 + 23)}  & 92.0$_{\pm 102.3}$  \\\hline
\end{tabular}
\caption{Number of documents for the six media types and their average number of tokens plus standard deviation (combing the train and test set).}
\label{tab:metadata}
\end{table}

Table~\ref{tab:metadata} shows the average number of documents for each one of the six media forms in our data and their average document length (including both train and test set) measured in the number of tokens. Distribution wise, the related statistics in the test set is similar.

\section{Experimental details}
\label{app:experiments}

\subsection{FNN details}

We use a simple FNN architecture with three hidden layers whose dimensionalities are 90, 60, 30, respectively. Each layer is a fully connected layer that consists of two linear transformations with a tanh activation function in between:

$$\mathrm{FNN\_layer}(x) = \tanh (xW_{1} + b_{1})W_{2} + b_{2}$$

We apply Adam optimizer with 5e-4 learning rate and 1e-5 L2
weight decay rate. We randomly take out 20\% of data from the train set to obtain a validation set, which is used for the five runs.  We train the model for 500 epochs and deploy the model with the best F1 on the held out validation set to the test set for evaluation.

\subsection{Linguistic features from \citet{zwthesis}}

According to \citet{zwthesis}, 2/3 of the 95 features come from \citet{Biber2006} with 42 of them also available in \citet{Biber1988}. 

These 95 features can be broken down into four categories: (1) structural features, such as mean word length, type-token ratio; (2) conversational features, such as contraction (e.g., ``I am'' $\rightarrow$ ``I'm''); (3) sentential features, which involve features related to passive voice, tense, coordination, and WH structure etc.; (4) lexical features, including part of speech, noun sub-categories, verb sub-categories, stance-related expressions, and so on. For full details, please refer to \citet{zwthesis}.

\subsection{XGBoost details}

We use the default configuration of the xgboost (v1.7.3) package in Python\footnote{\url{https://xgboost.readthedocs.io/en/stable/python/python_api.html}} for training the XGBoost classifiers, except for the ``max\_depth'' parameter, which we simply make equal to the current run number (i.e., 1, 2, 3, 4, 5).  


\subsection{Clustering details}

We use the best-performing pretrained SBERT model ``all-mpnet-base-v2''\footnote{\url{https://www.sbert.net/docs/pretrained_models.html}} to embed each text before clustering. For each clustering setup, we run the same experiments for three times to obtain small variations in the clustering results.

For KMEANS\footnote{\url{https://scikit-learn.org/stable/modules/generated/sklearn.cluster.KMeans.html}}, we vary the number of clusters (i.e., the \emph{k}) and use the following numbers: 10, 20, 30, 40, 50, 60, 70, 80, 90, 100, 150, 200, 250, 300, 500. This results in 45 ($= 15 \times 3$) different experiments.

For HDBSCAN\footnote{\url{https://hdbscan.readthedocs.io/en/latest/}}, we vary two paremters. One is the minimum cluster size, which is part of the HDBSCAN algorithm. The other is the dimensionality of the reduced SBERT embedding by UMAP, which is not part of the HDBSCAN algorithm, but essential for HDBSCAN to produce meaningful number of clusters. We set the following minimum cluster sizes: 10, 20, 40, 80, 100, 150, 200, 300, 400, 500. The size of reduced dimensionalities are 10, 30, or 50. This results in a total of $10 \times 3 \times 3 = 90$ different experiments.

The choices of parameters are not totally random, since some of them are somehow informed by our initial experiments. But they are not cherry picked either, since we simply use a wide range of numbers to vary the related parameters, without knowing the final results. 

As discussed in the paper, the main purpose for different clustering experiments is to aggregate them, either as a means of data augmentation or enhance model performance on classifying documents at the final stage of the pipeline.

\section{Results}
\label{app:results}

\begin{figure}
    \centering
    \includegraphics[width=1\columnwidth]{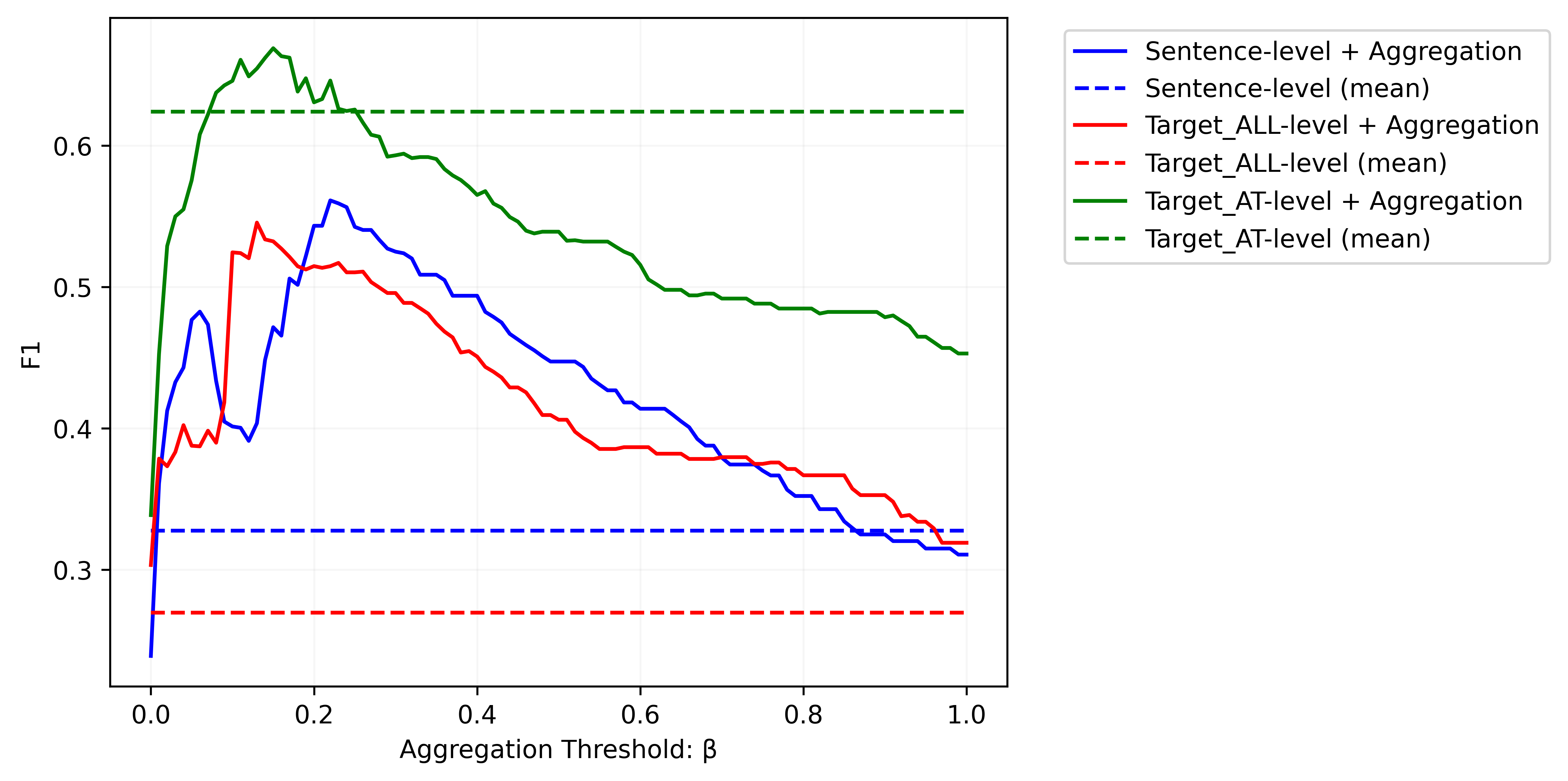}
    \caption{Aggregation versus no Aggregation with FNN as the high-influence cluster classifier. Results are averaged over the five runs.}
    \label{fig:aggregationFNN}
\end{figure}

Fig~\ref{fig:aggregationFNN} shows the performance variation of our models at different document part levels plus aggregation, as a function of the threshold $\beta$: the minimum number of times a document must be associated with a high-influence cluster in order to qualify as high-influence document, proportional to the total number of high-influence clusters available. The mean performance of these models on clusters from each clustering experiment is shown in dashed lines, as a baseline comparison.

\end{document}